# Automated Extraction of Fluoropyrimidine Treatment and Treatment-Related Toxicities from Clinical Notes Using Natural Language Processing

Xizhi Wu, MS[1], Madeline S. Kreider, PharmD, PhD[2], Philip E. Empey, PharmD, PhD[2], Chenyu Li, MS[1,3], Yanshan Wang, PhD[1,3,4,5*]

[1]Department of Health Information Management, University of Pittsburgh, Pittsburgh, PA, USA;

[2]Department of Pharmacy & Therapeutics, University of Pittsburgh, Pittsburgh, PA, USA;

[3]Department of Biomedical Informatics, University of Pittsburgh, Pittsburgh, PA, USA;

[4]Intelligent Systems Program, University of Pittsburgh, Pittsburgh, PA, USA; [5]Clinical and Translational Science Institute, University of Pittsburgh, Pittsburgh, PA, USA

*Corresponding author: Yanshan Wang, PhD, FAMIA; yanshan.wang@pitt.edu

**Abstract**

**Objective:** Fluoropyrimidines are widely prescribed for colorectal and breast cancers, but are associated with toxicities such as hand-foot syndrome and cardiotoxicity. Since toxicity documentation is often embedded in clinical notes, we aimed to develop and evaluate natural language processing (NLP) methods to extract treatment and toxicity information.

**Materials and Methods:** We constructed a gold-standard dataset of 236 clinical notes from 204,165 adult oncology patients. Domain experts annotated categories related to treatment regimens and toxicities. We developed rule-based, machine learning-based (Random Forest, Support Vector Machine [SVM], Logistic Regression [LR]), deep learning-based (BERT, ClinicalBERT), and large language models (LLM)-based NLP approaches (zero-shot and error-analysis prompting). Models used an 80:20 train-test split.

**Results:** Sufficient data existed to train and evaluate 5 annotated categories. Error-analysis prompting achieved optimal precision, recall, and F1 scores (F1=1.000) for treatment and toxicities extraction, whereas zero-shot prompting reached F1=1.000 for treatment and F1=0.876 for toxicities extraction.LR and SVM ranked second for toxicities (F1=0.937). Deep learning underperformed, with BERT (F1=0.873 treatment; F1= 0.839 toxicities) and ClinicalBERT (F1=0.873 treatment; F1 = 0.886 toxicities). Rule-based methods served as our baseline with F1 scores of 0.857 in treatment and 0.858 in toxicities.

**Discussion:** LMM-based approaches outperformed all others, followed by machine learning methods. Machine and deep learning approaches were limited by small training data and showed limited generalizability, particularly for rare categories.

**Conclusion:** LLM-based NLP most effectively extracted fluoropyrimidine treatment and toxicity information from clinical notes, and has strong potential to support oncology research and pharmacovigilance.

**Background and Significance**

Fluoropyrimidines (FPs), including capecitabine and 5-fluorouracil (5-FU), are chemotherapy agents often used in the treatment of colorectal and breast cancers [1]. However, FP use is associated with multiple adverse events, including hand-foot syndrome (HFS), cardiotoxicity and gastrointestinal toxicities [2]. One of the most common adverse drug reactions (ADR) for patients taking capecitabine is HFS, occurring in up to 42% of breast cancer patients and 53.5% of colorectal cancer patients [1], and a median onset of 79 days [2]. HFS is a dermatologic reaction affecting the skin of the palms of the hands and soles of the feet. HFS presents as painful redness, swelling, tingling, and in more severe cases blistering, peeling, or thickening of the skin [3]. Cardiotoxicity is a serious but rare toxic ADR [2] that occurs in approximately 5% of patients receiving a FP [4] [5]. It presents with chest pain, palpitations, dyspnoea, hypotension, arrhythmias, and in rare cases may lead to myocardial infarction, heart failure, or sudden cardiac death [5]. These toxicities can significantly impair patients' quality of life, clinical outcomes, and treatment course, often leading to medication interruptions, discontinuations, and dose or frequency decreases [6,7]. Large-scale and accurate identification of these toxicities in EHRs is critical for advancing clinical research and ultimately informing strategies for better toxicity prediction, prevention, and treatment.

EHRs are valuable sources to find and extract FP treatment and treatment-related toxicity information. ADRs are often documented in unstructured clinical notes rather than structured fields [8]. There are 3 commonly used methods in extracting such data from free-text clinical notes: manual chart review, International Classification of Diseases(ICD) codes, and NLP methods.

Manual chart review is a retrospective examination of individual patient medical records by trained reviewers to identify FP toxicities. This method allows researchers to capture ADRs, dose modifications, or reasons for treatment discontinuation [9–11], as documented in physician notes or other unstructured sections of the EHR. Many studies have relied on manual review of free-text notes to extract FP toxicity information. Jiang et al.[12] assess how older patients tolerate capecitabine by having two trained annotators analyzing toxicities such as HFS, fatigue and diarrhea from EHRs and patient reported outcomes (PRO). Hennessy et al.[9] conducted a manual review of oncology patients' records at M.D. Anderson Cancer Center to extract detailed toxicity data associated with different starting doses of capecitabine. Van Beek et al.[13] conducted a retrospective review of EHRs from the oncology department of Radboud University Medical Center. Capecitabine toxicities were manually identified and graded using the NCI CTCAE v3.0 (Common Terminology Criteria for Adverse Events) [14]. While manual chart review provides high clinical fidelity, it is time-consuming [15], resource-intensive, and subject to documentation variability and inter-rater inconsistency [16]. Therefore, we propose NLP methods to reduce the effort and time needed for identifying FP treatment and treatment-related toxicities from unstructured clinical text.

Structured diagnosis codes, such as the International Classification of Diseases (ICD), are commonly used in claims and administrative data to identify ADRs. For example, Nishijima et al.[17] utilized ICD-9 codes from Medicare records to extract ADRs including diarrhea, dehydration, and cardiotoxicity in older adults treated with capecitabine. Bhimani et al.[17] extracted capecitabine-related toxicities from EHR-derived structured data using predefined ICD-9/10 diagnosis codes, laboratory values, and treatment timelines. Li et al.[18] identified cardiotoxicity events using predefined ICD-9-CM diagnosis codes for cardiac conditions such as heart failure, arrhythmia, and myocardial infarction, treating these events as indicators of chemotherapy-induced cardiotoxicity. However, reliance on billing codes may lead to underreporting or misclassification[19][20], particularly for lower-grade or undocumented toxicities that are not formally coded. While ICD-based extraction provides scalability and standardization, it lacks the detail available from the clinical notes within EHR to appropriately identify patients with and without the toxicity [21]. Our study focuses on extracting FP Treatment and treatment-related toxicities from unstructured EHR narratives. Future work may involve integrating NLP-based approach with structured data sources, such as ICD-coded diagnoses, to improve coverage and accuracy.

Natural language processing (NLP) is a technique widely used to extract ADRs from unstructured EHR data [22]. NLP is a field of artificial intelligence focused on enabling computers to understand, interpret, and generate human language [23]. In the clinical domain, NLP techniques are applied to extract meaningful information from unstructured text, such as physician notes, discharge summaries, and progress reports [24]. These texts often contain rich clinical narratives, including detailed descriptions of symptoms, adverse events, and clinical decisions that are not recorded in objective, structured data fields. In terms of extracting ADRs, NLP methods include a range of machine learning techniques, including neural sequence labeling models like Bidirectional Long Short-Term Memory Networks with a Conditional Random Field layer[25] for named entity recognition and models such as attention-based bidirectional Long Short-Term Memory Networks[26], SVM[27], and RF[28] for relation classification[29]. Hong et al.[30] developed an NLP pipeline based on Apache Mayo clinical Text Analysis and Knowledge Extraction System (cTAKES) [31] to automatically extract CTCAE [32] v5.0-defined toxicities related to capecitabine and other chemotherapies. The NLP pipeline extracted from free-text oncology treatment notes using Unified Medical Language System(UMLS)-based Named Entity Recognition (NER), assertion detection, and post-processing rules. The NLP pipeline by Hong was validated against expert annotation and achieved a precision of 0.91, recall of 0.72, and F1 score of 0.80 in extracting CTCAE v5.0 toxicities. Even though only one study targeting extracting capecitabine toxicity is found, there are a lot of studies that utilize NLP methods to extract adverse drug events with other cancer drugs. For example, Fan et al.[33] developed a BERT based deep learning model to extract ADEs from open health forums (WebMD, Drugs.com). Although not specific to capecitabine, this study included a wide variety of anticancer drugs and demonstrated state-of-the-art performance (F1 = 0.97 for ADE extraction). Li et al.[34] developed an end-to-end deep learning model trained on the MADE 1.0 dataset[29], a publicly available corpus of expert-annotated electronic health records from oncology patients, specifically designed for extracting medication, indication, and ADE information. Their system combined a BiLSTM-CRF for named entity recognition and a BiLSTM-Attention mechanism for relation extraction, enabling high-quality extraction of medication-related events from clinical narratives. Our approach offers broader coverage than previous works, as we systematically compared the performance of rule-, machine learning, deep learning, and LLM based methods within the same framework. Notably, to our knowledge, prior studies have not incorporated LLM-based methods for FP toxicity extraction, making our work the first to assess their potential in this context.

In this paper, our objective is to define, develop and evaluate multiple NLP systems—rule-, machine learning, deep learning and LLM based systems—for capturing FP treatment and treatment-related toxicities from EHR clinical notes, and to identify the most effective approach. We make four contributions: (1) curate an annotated gold-standard corpus of 236 oncology notes; (2) perform a comparison of the performance of four NLP systems; (3) first to apply the error-analysis prompting method to capture FP treatment and treatment-related toxicities and achieved the highest overall performance; (4) provide error analyses on 5 category and practical design to inform NLP pipelines for capturing FP treatment and treatment-related toxicities from unstructured EHR text.

**Materials and Methods**

Figure 1 illustrates the entire workflow for the extraction of FP treatment and treatment-related toxicities. In the following subsections, we detail the data collection, FP treatment and treatment-related toxicity categories, and NLP approaches.

Figure 1. Workflow for the Extraction of Fluoropyrimidine Treatment and

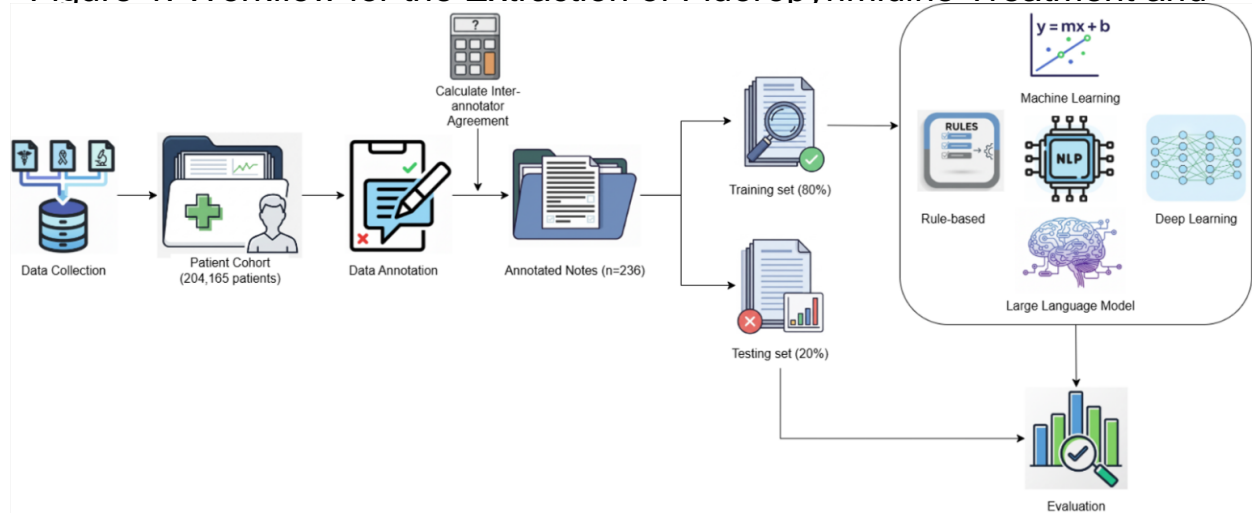

*Data Collection*

We first identified a cohort of adult patients diagnosed with breast or gastrointestinal cancer between January 1, 2016 and December 31, 2022 at UPMC through ICD-10 codes. Then patients with an EHR medication order for a fluoropyridine (i.e., capecitabine, 5-FU) from the UPMC health system between January 1, 2026 and December 31, 2023 were included for downstream analysis. All de-identified clinical notes during this timeframe were aggregated and disseminated through the University of Pittsburgh Clinical and Translational Science Institute Health Record Research Request (R3). These notes were exported from multiple types of clinical documents, including discharge summaries, outpatient progress notes, and physician reports. The clinical notes selected for inclusion in the corpus were aligned with each patient's capecitabine initiation date, ensuring that all notes analyzed were temporally relevant to the course of capecitabine therapy and its potential treatment-related toxicities. This study was reviewed and designated as exempt by the University of Pittsburgh Institutional Review Board.

*Fluoropyrimidine Treatment and Toxicity Category and Annotation*

A multidisciplinary team of clinical experts including a hematologist/oncologist, a cardio-oncologist, an oncology pharmacist, and a pharmacist clinical researcher served as domain experts to develop 35 term categories covering FP treatment information, cardiotoxicity, and HFS. Initial term lists were assembled using both expert input and the Common Terminology Criteria for Adverse Events (CTCAE) v.5. Categories were then systemically expanded through the National Cancer Institute's (NCI) Enterprise Vocabulary Services (EVS) Explore tool and the National Library of Medicine Unified Medical Language System (UMLS) to incorporate synonyms and related terminology [35]. All the 35 term categories are included in the Supplementary Materials.

To create our gold standard dataset, we randomly sampled 300 clinical notes from the eligible patient cohort. During the annotation process, 64 notes were excluded as ineligible due to insufficient content, ambiguous documentation, or lack of relevant FP treatment information, resulting in 236 annotated notes for model development and evaluation. The domain experts independently annotated an initial set of 36 notes to establish inter-annotator agreement, achieving an average pairwise Cohen's kappa of 0.77, indicating substantial agreement. Following agreement calibration, each expert annotated 50 additional notes, contributing to the final gold standard dataset of 236 notes. This annotation workflow is illustrated in Figure 1.The 236 eligible notes were split 80:20 for training and testing for each category. This sample size meets established criteria for binary classification: (1) exceeds the minimum 10 events per variable for stable logistic regression, (2) provides >80% statistical power to detect meaningful performance differences ($\Delta$ F1=0.15) between methods, and (3) demonstrates convergent learning curves and stable cross-validation performance (SD<0.05). The binary nature of our classification task and focused feature set allow for robust model development within this sample size.

Five annotation categories had sufficient representation and were defined to represent FP treatment and toxicities. FP treatments were annotated under the category drug of interest, which included capecitabine and 5-fluorouracil, their brand names (e.g., Xeloda, Adrucil) and abbreviations including either FP (e.g., 5-FU, FOLFOX, CAPEOX). FP treatment-related toxicities were defined as conditions observed in the context of FP use and likely attributed to the medication. These were annotated under the categories of arrhythmia, HF, valvular complications and HFS treatment/prevention therapies. Arrhythmia is defined as an abnormal rhythm of the heart [36], and its category captures mentions of cardiac rhythm disturbances. HF is a cardiotoxicity side effect of FPs [37] and its category captures mentions of HF (e.g., heart failure, congestive heart failure) and related conditions (e.g., edema, leg swelling). Valvular complications refer to abnormalities or dysfunctions of the heart valves [38] and its category capture structural abnormalities (e.g., bicuspid aortic valve) and functional abnormalities (e.g., tricuspid regurgitation). HFS treatment/prevention therapies [3] refers to treatment and preventive medications used for HFS and its category captures mentions of prescription and over-the-counter topicals used to reduce HFS incidence or severity. Their definition and keywords are listed in **Table 1**.

**Table 1**. FP Treatment and Treatment-related Toxicity

|  | Category | Definition | Keywords |
|--|----------|------------|----------|
|  |          |            |          |

| Fluoropyrimidine Treatment | Drug of interest | Mentions of capecitabine or 5-FU, including brand names, combination therapies and abbreviations | Capecitabine, Xeloda, Xitabin 5-FU, 5-Fluorouracil, Fluoro Uracil, Adrucil, Carac, Flurablastin CAPOX, CAPIRI, CAPEOX, CAPEMONO, FOLFOX, FOLFIRI, FOLFOXIRI, MFOLFOX, AIO, De Gramont Regimen, XELOX, XELIRI, FOLFIRINOX. |
|---|---|---|---|
| Treatment-related Toxicity | Arrhythmia | An abnormal rhythm of the heart | Cardiac Arrhythmia, Dysrhythmia, Irregular Heartbeat, Heart Rhythm Disturbance, Cardiac Rhythm Disorder Afib, A Fib, Atrial Flutter, Auricular Flutter, A-flutter, AF, Auricular Fibrillation VF, Ventricular Fibrillation, Cardiac Arrest Due to VF, Ventricular Tachycardia (V TACH), Tachycardia;Ventricular Flutter Atrial, Fibrillation Atrial, Flutter Auricular, Heart Arrhythmia |
| | Heart failure | Mentions of heart failure and related conditions associated with capecitabine use. | HF, Cardiac Failure, Heart Insufficiency, Myocardial Failure, Cardiac Insufficiency, Bilateral Leg Edema, Swelling, Dropsy, Hydrops, Oedema, Fluid Overload Reduced Ejection Fraction (EF or LVEF), Reduced LV Function Cardiogenic |

|  |  |  | Shock, Heart Shock, Cardiovascular Collapse, HF Exacerbation. |
|  | Valvular complications | The structural abnormalities and functional abnormalities of the heart valves associated with capecitabine use. | TR, Tricuspid Insufficiency, Incompetence, Right AV Valve Regurgitation AR, Aortic Regurgitation, Aortic Incompetence, Aortic Valve Insufficiency Valve Disorder, AV Valve Abnormality, Valvular Dysfunction. |
|  | HFS treatment/prevention therapies | The mentions of topical measures used to reduce HFS severity or incidence | Cream, Urea Cream, Aqua Care, Nutraplus, Vanamide, Carbamide, Elaqua XX, Lanaphilic, Ureaphil, Carbamide, Utterly Smooth, Udderly Smooth Cream, Lotion, Gel, Ointment, Salve, Solution, Suspension Uridine Triacetate (Vistogard) |

***Rule-based NLP Algorithm***

We developed a rule-based NLP algorithm for the toxicities extraction using MedTagger, a clinical NLP tool based on the Unstructured Information Management Architecture (UIMA) framework. The MedTagger software is publicly available at GitHub (https://github.com/OHNLP/MedTagger). Clinical experts provided a list of keywords and synonyms for each toxicity concept from medical terminologies and ontologies. We then used 80% of the gold standard dataset as training data to develop regular expression rules for the NLP algorithm. MedTagger facilitated the execution of these regular expression-based rules, allowing the algorithm to annotate and extract information from unstructured clinical text data. We applied an 80:20 train-test split across all toxicity categories, and this split was kept consistent across all methods. We specified customized negation rules based on patterns observed in our training data. These context rules handle various linguistic patterns including negation markers, uncertainty indicators, historical references, and experiencer identification. The regular expression patterns for toxicity concepts were developed to capture drug of interest, arrhythmia, HF, valvular complications, and HFS treatment/prevention therapies. The rule-based NLP algorithm including the customized negation rules can be found in the Supplemental Materials.

*Machine Learning Algorithms*

We implement a machine learning pipeline for binary classification where separate models are trained for each toxicity category. This design choice allows for independent optimization of each toxicity type and better handling of class imbalance issues that are common in medical datasets.

The pipeline contains three different machine learning algorithms for comparison: RF, SVM and LR. The pipeline follows a four-step workflow. First, data preprocessing is performed, which involves tokenizing sentences, converting text to lowercase, and removing non-informative symbols. Second, vectorization is used to transform text into numerical features for model input. Linear models (SVM, LR) use term frequency–inverse document frequency (TF-IDF)[39] features with sublinear term scaling. Ensemble models (RF) use count-based n-gram features[40], which allow tree-based methods to exploit raw term frequencies without the normalization imposed by TF-IDF. Third, model training is conducted with class weighting set to balanced to address label imbalance. Fourth, validation is performed on test set and report precision, recall, F1 score for downstream comparison. We process five specific toxicity categories: drug of interest, arrhythmia, HF, valvular complications, and HFS treatment/prevention therapies. For each category, one binary classifier was trained, resulting in 15 total models (5 categories × 3 algorithms). Each model learns to distinguish between sentences that contain information about its specific toxicity type.

*Deep Learning Algorithms*

We implement a deep learning pipeline based on two models: BERT[41] and ClinicalBERT [42]. The deep learning pipeline follows a four step workflow. First, we tokenize the input text using BERT vector tokenizer. Second, contextual embeddings are generated by passing the tokenized sequences through either BERT model or ClinicalBERT, with the latter pre-trained on biomedical and clinical text to better capture domain-specific terminology. Third, model training is conducted independently for each toxicity category to enable task-specific adaptation. Fourth, evaluation is performed on a testing set, converting probabilistic outputs to binary predictions, and reporting precision, recall, and F1 score.

*Large Language Model-based Approach*

We implemented two prompting strategies using LLaMA 3.1 8B: zero-shot prompting [39] and error analysis prompting[43] .

**Zero-shot Prompting**: Zero-shot prompting performs binary classification without training examples. Each prompt contains three components: (1) the classification task description, (2) domain-specific medical terminology lists, and (3) explicit instructions for binary ("yes"/"no") responses with explanations. For example, the HFS treatment/prevention prompt includes comprehensive lists of topical agents, while the arrhythmia prompt contains rhythm disturbance terminology. **Table 2** shows the Zero-shot prompt used in this study.

**Table 2**. Example of Zero-shot Prompt and Error Analysis Prompt on Heart Failure

| Zero-shot Prompt | You are given a sentence from a clinical text, if that sentence contains any information related to instances of HF, respond with yes and explain why. If not, respond with no and explain why. These words being the signs and evidence: HF, cardiac failure, heart insufficiency, myocardial failure, cardiac insufficiency bilateral leg edema, swelling, dropsy, hydrops, oedema, fluid overload reduced ejection fraction (EF or LVEF), reduced LV function cardiogenic shock, heart shock, cardiovascular collapse, HF exacerbation. If these words are mentioned in the sentence, respond with yes and explain why. If these words are not mentioned in the sentence, respond with no and explain why. |
|---|---|
| Error Analysis Prompt | You are given a sentence from a clinical text, if that sentence contains any information related to instances of HF, respond with yes and explain why. If not, respond with no and explain why. These words being the signs and evidence:: HF, cardiac failure, heart insufficiency, myocardial failure, cardiac insufficiency bilateral leg edema, swelling, dropsy, hydrops, oedema, fluid overload reduced ejection fraction (EF or |

|  | LVEF), reduced LV function cardiogenic shock, heart shock, cardiovascular collapse, HF exacerbation. If these words are mentioned in the sentence, respond with yes and explain why. If these words are not mentioned in the sentence, respond with no and explain why. Here is a sentence: Trace edema bilateral lower extremities. Reasoning: Step 1: Read the sentence and identify key terms that may indicate signs or evidence of HF." Step 2: The phrase "bilateral lower extremities edema" directly matches one of the listed indicators: "bilateral leg edema", which is a known clinical sign of fluid overload. Step 3: Although the sentence does not explicitly mention a diagnosis of HF, bilateral edema is a common and recognized symptom associated with congestive heart failure (CHF). Step 4: Since the sentence includes a relevant physical finding from the keyword list, it provides indirect evidence consistent with possible HF. Answer: Yes. The sentence mentions "bilateral lower extremities edema," which matches "bilateral leg edema," a recognized sign suggestive of HF or fluid overload. |
|--|--|

**Error Analysis Prompting:** Error analysis prompting enhances zero-shot performance by incorporating chain-of-thought (CoT) reasoning[44] derived from systematic error analysis. Figure 2 illustrates the four-step process. 1) Error Identification - We applied zero-shot prompting to the training set and systematically cataloged misclassifications, identifying recurring error patterns for each toxicity category. 2) Prompt Enhancement - False positive and false negative cases from Step 1 were analyzed to create corrective examples. The enhanced prompt combines: (a) original zero-shot instructions, and (b) CoT reasoning examples that demonstrate correct classification of previously misclassified cases. 3) Test Set Application - The error-analysis enhanced prompts were applied to the held-out test set for final classification. 4) Performance Evaluation - Model outputs were evaluated against gold-standard annotations to calculate performance metrics.

Each CoT reasoning example follows a structured four-part analysis: (1) systematic parsing of clinical text, (2) identification of medical concepts, (3) matching against reference terminology, and (4) evidence-based classification decision. **Table 2** provides complete examples of both prompting approaches.

To illustrate this process (as shown in **Figure 2**), consider heart failure detection. The zero-shot prompt initially misclassified "bilateral lower extremity edema" as negative. Through error analysis (Step 1 in Figure 2), we identified this pattern and created a CoT example (Step 2 in Figure 2) demonstrating that bilateral edema represents fluid overload, a key heart failure indicator. This corrective reasoning was incorporated into the enhanced prompt, teaching the model to recognize indirect clinical manifestations.

The error analysis approach systematically addresses category-specific challenges. For arrhythmia, examples demonstrate distinguishing rhythm disturbances from general cardiac conditions. For valvular complications, examples clarify the difference between functional and structural abnormalities. This targeted error correction enables the model to handle complex clinical language patterns identified through training set analysis.

Figure 2. Error Analysis Prompting LLM.

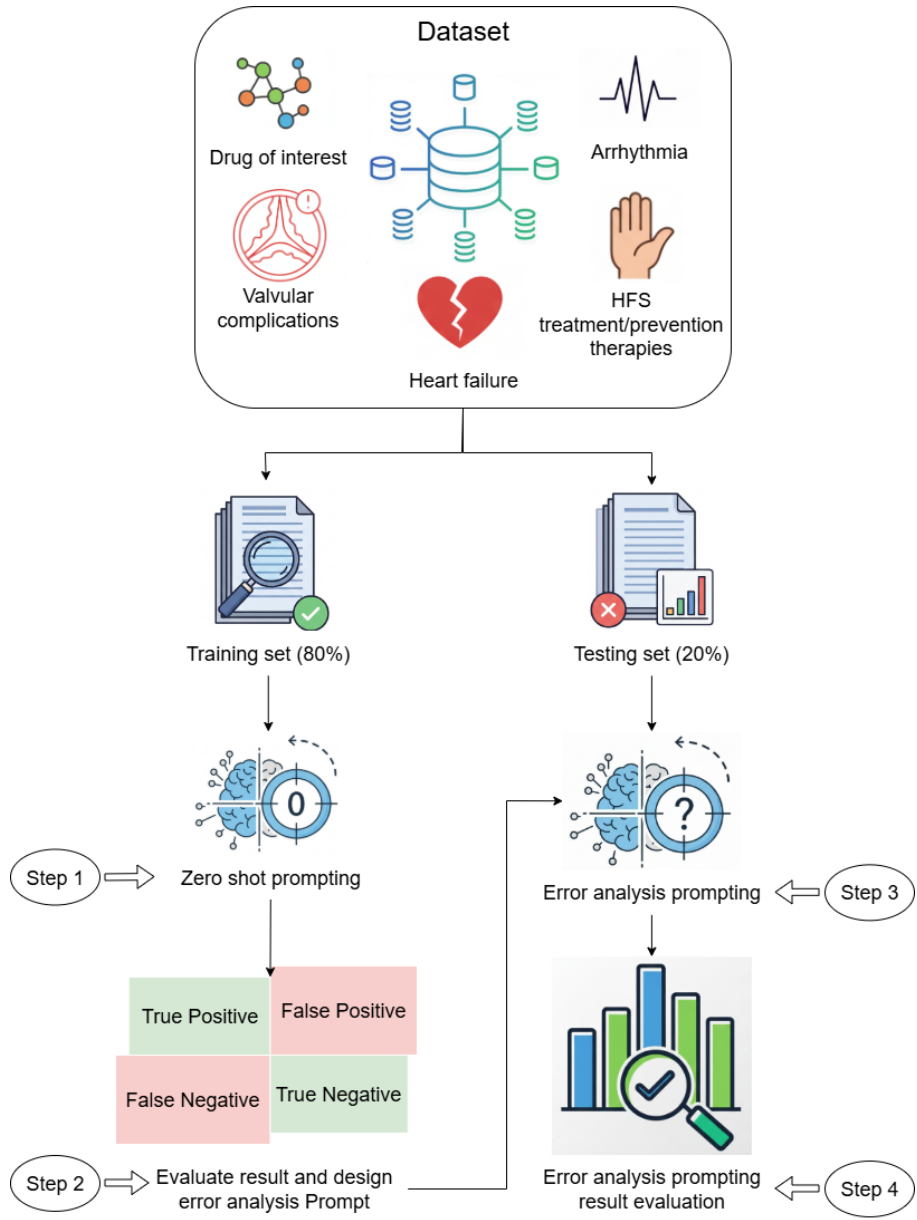

*Evaluation*

We use the standard performance metrics, including precision, recall, and F1 score, as defined below. We use weighted F1 scores as the toxicity categories are imbalanced.

$$Precision = \frac{True\ Positive}{True\ Positive\ +\ False\ Positive}$$

$$Recall = \frac{True\ Positive}{True\ Positive\ +\ False\ Negative}$$

$$F1\ score = \frac{2\ \times\ Precision\ \times\ Recall}{Precision\ +\ Recall}$$

**Results**

The performance of the rule-based NLP algorithm, machine learning, and LLMs in extracting FP treatment and treatment-related toxicity is presented in **Table 3**. Among LLM-based NLP algorithms, the error analysis prompting approach using LLaMA3 8b achieved the optimal overall performance with a F1 score of 1.000 in FP treatment and treatment-related toxicity. This result demonstrates the effective generalization capabilities of large language models (LLMs) when provided with task-specific examples. The model was able to match expert-level annotation, suggesting promising potential for practical clinical NLP applications. The zero-shot prompting method using LLaMA3 8b also showed robust performance, particularly in the FP treatment extraction, where it reached F1 score of 1.000 in drug-of-interest category. While it did not match the error analysis prompting optimal scores in treatment-related toxicity extraction, its performance was still high in Arrhythmia (F1 = 1.000), Valvular Complications (F1 = 0.886) and HFS treatment/prevention therapies (F1 = 0.920), further validating the utility of LLMs prompting even without fine-tuning or in-context examples. However, the zero-shot prompting approach did not do well for HF (F1 = 0.696).

**Table 3**. Performance of the rule-, machine learning-, deep learning- and LLM- based algorithms on Extraction of FP Treatment and Treatment-related Toxicity on test set.

|  | Treatment | Treatment-related Toxicity |
|--|--|--|

| Precision Recall F1 score | Capecitabine (n=8) | Arrhythmia (n=14) | Heart Failure (n=11) | Valvular Complications (n=9) | HFS treatment/ prevention therapies (n=25) | Average |
|---|---|---|---|---|---|---|
| Rule-based NLP | 1.000 | 1.000 | 0.800 | **1.000** | 1.000 | 0.950 |
| | 0.750 | 0.714 | 0.800 | **1.000** | 0.667 | 0.794 |
| | 0.857 | 0.833 | 0.800 | **1.000** | 0.800 | 0.858 |
| SVM | 0.900 | 0.938 | 0.818 | **1.000** | 1.000 | 0.939 |
| | 0.875 | 0.929 | 0.818 | **1.000** | 1.000 | 0.937 |
| | 0.873 | 0.928 | 0.818 | **1.000** | 1.000 | 0.937 |
| LR | 0.900 | 0.938 | 0.818 | **1.000** | 1.000 | 0.939 |
| | 0.875 | 0.929 | 0.818 | **1.000** | 1.000 | 0.937 |
| | 0.873 | 0.928 | 0.818 | **1.000** | 1.000 | 0.937 |
| RF | 0.786 | 0.889 | 0.731 | 0.907 | 0.963 | 0.873 |
| | 0.625 | 0.857 | 0.727 | 0.889 | 0.960 | 0.858 |
| | 0.564 | 0.854 | 0.723 | 0.886 | 0.960 | 0.856 |
| Bert | 0.900 | 0.889 | 0.731 | 0.778 | **1.000** | 0.850 |
| | 0.875 | 0.857 | 0.727 | 0.778 | **1.000** | 0.841 |
| | 0.873 | 0.854 | 0.723 | 0.778 | **1.000** | 0.839 |
| ClinicalBert | 0.900 | **1.000** | 0.818 | 0.907 | 0.963 | 0.922 |
| | 0.875 | **1.000** | 0.727 | 0.889 | 0.960 | 0.894 |
|   | 0.873 | **1.000** | 0.696 | 0.886 | 0.960 | 0.886 |
| Zero-Shot Prompting | **1.000** | **1.000** | 0.818 | 0.907 | 0.931 | 0.914 |
|   | **1.000** | **1.000** | 0.727 | 0.889 | 0.920 | 0.884 |
|   | **1.000** | **1.000** | 0.696 | 0.886 | 0.920 | 0.876 |
| Error Analysis Prompting | **1.000** | **1.000** | **1.000** | **1.000** | **1.000** | **1** |
|   | **1.000** | **1.000** | **1.000** | **1.000** | **1.000** | **1** |
|   | **1.000** | **1.000** | **1.000** | **1.000** | **1.000** | **1** |

For deep learning-based NLP algorithms leveraging pre-trained language models, ClinicalBERT outperformed the base BERT model in extracting a few treatment-related toxicities, such as Arrhythmia (F1 = 1.000) and Valvular Complications (F1 = 0.886). Both BERT-based models reached an F1 score of 0.873 for FP Treatment extraction in the drug-of-interest category and performed similarly in the HFS treatment/prevention therapies category (ClinicalBert F1 = 0.960 and Bert F1=1.000). However, both models struggled with Heart Failure, where F1 scores ranged between 0.696 and 0.723. Bert model yields the lowest average performance in treatment-related toxicities extraction (F1 = 0.839).This suggests that certain categories may require additional domain-specific adaptation for BERT-based models.

Among traditional approaches, the rule-based NLP system served as our baseline, achieving a F1 score of 0.857 in FP Treatment extraction and an average F1 score of 0.858 in treatment-related toxicities extraction. It performed particularly well onValvular Complications (F1 = 1.000), outperforming SVM, RF and deep learning methods on these categories. This suggests that when domain knowledge is available, rule-based systems can remain competitive with machine learning and deep learning methods. Machine learning models like LR and SVM also delivered robust results, each reaching an average F1 score of 0.937 in treatment-related toxicities extraction. It is noteworthy that LR and SVM yielded identical results, which may be attributed to the small validation set and the similarity of the two algorithms. Both SVM (with a linear kernel) and logistic regression are linear classifiers that aim to identify optimal separating hyperplanes. On small, linearly separable datasets, they frequently converge to similar solutions. These models performed best in categories such as HFS treatment/prevention therapies and Valvular Complications, both achieving perfect F1 scores of 1.000. However, performance on HF was noticeably lower (F1 = 0.818), which may be attributed to class imbalance and variability in language in the dataset.

Lastly, the RF classifier yielded the lowest performance in extracting FP treatment with a F1 score of 0.564. Among treatment-related toxicities, it struggled particularly in identifying HF (F1 = 0.723), indicating that tree-based models may lack the flexibility to capture complex patterns in clinical narratives without substantial feature engineering.

**Discussion**

In this study, we compared rule-based, machine learning, deep learning, and LLM–based methods for extracting FP treatment and treatment-related toxicities from oncology clinical notes. Error analysis prompting approach achieved the highest overall performance, indicating that when guided by carefully constructed prompts incorporating representative error cases, LLMs can match expert-level annotation even in specialized clinical domains. Zero-shot prompting also performed strongly, underscoring the potential for rapid deployment without fine-tuning. Rule-based methods performed competitively, particularly for the valvular complications category, reflecting their continued value in settings with limited training data. Traditional machine learning and deep learning models were effective for most categories in treatment-related toxicity, but they struggled with the Heart Failure category. This is because many heart failure findings are indirect (e.g. leg edema), highlighting the challenge of extracting conditions expressed through indirect clinical findings.

The toxicity categories in our study show strong conceptual alignment with those in Hong et al[30], as both were anchored to CTCAE v5.0 terminology (e.g., arrhythmia, heart failure). This overlap supports the comparability of our results and enhances the validity of our findings

This study has several limitations. First, the dataset is small (236 annotated notes) and drawn from a single institution, which may limit generalizability to other healthcare settings or EHR systems. Second, we focused on a single drug class (FPs) and a subset of toxicity categories, which may not fully represent the complexity of oncology drug toxicities. Third, our final terminology set was constrained by the initial development corpus, which limited the discovery of additional clinically relevant toxicity terms and may restrict model prediction and generalizability.

**Conclusion**

In this study, we developed and compared rule-, machine learning-, deep learning-, and large language models (LLM)-based NLP systems for FP treatment and treatment-related toxicities extraction from oncology clinical notes. Our findings demonstrate that error analysis prompting with LLM achieved the highest overall performance. Error analysis prompting and zero-shot prompting achieved the best performance in FP Treatment extraction. Machine Learning methods achieved second best performance in treatment-related toxicities extraction. Deep learning based methods underperformed and were most challenged by heart failure categories. These findings suggest that carefully designed prompting can enable LLM to match expert-level annotation. Our results demonstrate the promise of LLM-based prompting-enhanced by error-analyzed examples-for extraction of FP treatment and treatment-related toxicities from oncology clinical notes.

**Future Work**

Future research should prioritize validating these methods in independent cohorts of patients receiving fluoropyrimidines for gastrointestinal and breast cancers. Multi-institutional validation across diverse healthcare systems would assess generalizability of model performance and robustness to variations in clinical documentation practices. Testing in different geographic regions and EHR platforms would further establish the reliability of our approach for fluoropyrimidine toxicity extraction.

Methodological improvements should focus on automated prompt optimization for LLMs and exploration of additional architectures to enhance interpretability and reliability. Integration of structured data elements could improve toxicity detection, though the utility varies by toxicity type. While cardiotoxicity manifestations have corresponding ICD-10 codes (e.g., I48 for atrial fibrillation, I50 for heart failure), medication-induced hand-foot syndrome lacks specific diagnostic codes, highlighting the particular importance of NLP extraction for toxicities without structured coding options. Combining NLP with available structured data—including laboratory values, vital signs, and relevant ICD codes where applicable—could enhance overall system performance.

Temporal modeling to capture the relationship between drug exposure and toxicity onset represents another important direction, as our current approach does not explicitly model time-dependent relationships. Development of a standardized fluoropyrimidine toxicity ontology would improve consistency across institutions and facilitate integration into clinical decision support systems[45]. Such an ontology could harmonize toxicity definitions, severity grading, and documentation standards, enabling more reliable cross-institutional research and real-world evidence generation.

Finally, evaluation in clinical research settings would assess the impact of this approach on advancing the understanding of treatment-related toxicities. Reliable and scalable extraction could improve characterization of adverse events, guide future studies on toxicity mechanisms, and inform strategies for treatment optimization and prevention in patients receiving FPs.

**Acknowledgements**

The authors would like to acknowledge the support of the University of Pittsburgh CTSI infrastructure, supported by the National Institutes of Health through Grant Number UL1TR001857. The authors also extend their gratitude to the clinical team members for their invaluable contributions: Anas Hanini, PharmD, MBA (formerly UPMC Shadyside, currently Cleveland Clinic Abu Dhabi, Abu Dhabi, UAE); Ashish Patel, MD (UPMC Shadyside, Pittsburgh, PA); Dominika Dulak, MD (UPMC Shadyside, Pittsburgh, PA); Joshua Levenson, MD (UPMC Shadyside, Pittsburgh, PA); Timothy Brenner, PharmD, BCOP (UPMC Shadyside, Pittsburgh, PA); and Melissa Bastacky, PharmD, BCOP (UPMC Shadyside, Pittsburgh, PA).

**Author contributions**

XW and MSK: conceptualized the study, conducted data analysis, wrote the manuscript; CL: conducted data analysis, wrote the manuscript; PEE and YW: conceptualized the study, wrote the manuscript.

**Funding**

This work was supported by the Shear Family Foundation.

**Conflicts of interest**

The authors declare that they have no conflicts of interest related to this publication.

**Data availability**

The data used in this study are electronic health records that may contain sensitive protected health information (PHI). Thus, it is not available for public access. The NLP algorithms are publicly available at GitHub: https://github.com/PittNAIL/NLP4FPandToxicity.

**Supplementary materials**

| Category | Keywords |
|---|---|
| Drug of interest | Capecitabine, Xeloda, Xitabin 5-FU, 5-Fluorouracil, Fluoro Uracil, Adrucil, Carac, Flurablastin CAPOX, CAPIRI, CAPEOX, CAPEMONO, FOLFOX, FOLFIRI, FOLFOXIRI, MFOLFOX, AIO, De Gramont Regimen, XELOX, XELIRI, FOLFIRINOX. |
| HFS treatment/prevention therapies | Cream, Urea Cream, Aqua Care, Nutraplus, Vanamide, Carbamide, Elaqua XX, Lanaphilic, Ureaphil, Carbamide, Utterly Smooth, Udderly Smooth Cream, Lotion, Gel, Ointment, Salve, Solution, Suspension Uridine Triacetate (Vistogard) |
| Treatment modifiers | Capecitabine induced, Fluoropyrimidine induced, FP induced, Xeloda induced, 5-FU induced, Chemo induced, Chemotherapy induced |
| Treatment doses | 14 ON, 7 ON, 14 OFF, 7 OFF, 1-14 DAYS ON, 1-14 ON, 1-14 DAYS OFF, 1-14 OFF, BID, twice daily, once daily, 1-2 time(s) a day |
| Radiation | Chemoradiation, Chest radiation |
| Treatment Outcomes | Dose decrease, decrease dose, discontinue, Discontinuation, delay, hold, dose reduction, reduce dose, reduce frequency, frequency reduction, stop, start, initiate, initiation, begin, began, started, day 1, cycle 1 |
| HFS | Acral Erythema, Chemotherapy-Induced, Acral Erythemas, Chemotherapy-Induced, Chemotherapy Induced Acral Erythema, Chemotherapy Induced Palmoplantar Erythrodysesthesia, Chemotherapy-Induced Acral Erythema, Chemotherapy-Induced Acral Erythemas, Chemotherapy-Induced Palmoplantar Erythrodysesthesia, Chemotherapy-Induced Palmoplantar Erythrodysesthesias, Hand and foot syndrome secondary to chemotherapy, Hand Foot Syndrome, Hand-Foot Syndrome, Hand-Foot Syndromes, Palmar-plantar erythrodysaesth., Palmar-plantar erythrodysaesthesia syndrome, palmar-plantar erythrodysesthesia, Palmar-plantar erythrodysesthesia syndrome, Palmar-Plantar Erythrodysthesia, Palmoplantar Erythrodysesthesia, Chemotherapy-Induced, Palmoplantar Erythrodysesthesias, Chemotherapy-Induced, Syndrome, Hand-Foot, Syndromes, Hand-Foot, palmar-plantar erythrodysesthesia syndrome (diagnosis), hand-foot, HFS, CiHFS, palmar plantar erythrodysthesia, PPE, palmar-plantar, hand and foot, hand and foot syndrome, L27.1, capecitabine-induced hand-foot syndrome |
| Skin Integrity | Peeling, Peeling of skin, Desquamation, Desquamative state, Scaling, Dropping of scales, Exfoliating, Flaking of skin, Shedding of scales, Exfoliation, Scaly skin, peeling skin, |
|  | flaking skin, Skin exfoliation, Skin peeling, Skin desquamation, Exfoliation of skin, Scaling of skin, Skin scaling, Scaling of the skin, skin appearance peeling, skin scales, scales, peeling skins, skin peel, exfoliate, peel, skin flaking, desquamation; skin, Peeling;skin, Desquamation of skin, Blisters, skin vesicle, Have Blistering, Has blistering, Blister of skin, Cutaneous blister, Dermal blister, Skin blister, Skin blisters, Skin blistering, Blistering skin, blister skin, Bleeding skin, bleeding of skin, Skin bleeding, bleed skin, bleeds skin, Fissures, Skin Fissure, Skin fissuring (palms), Skin fissuring (soles), Fissure in skin, Cracks in skin, Splits in skin, Cracked skin, Fissuring of skin, cracking of skin, skin crack |
| Edema | Edema of palm of hand, edema of left palm of hand, edema of bilateral palm of hand, edema of right palm of hand, Edema of sole of foot |
| Hyperkeratosis | Grade 1 Hyperkeratosis, Increased Keratinization, Grade 3 Hyperkeratosis, Excessive cornification, Keratosis, Tylosis, Hyperkeratoses, hyperkeratosis of skin, skin texture, hyperkeratosis, skin hyperkeratosis |
| Pain | Pain of left hand, pain in left hand only, Pain in right hand, pain in right hand only, limb pain right hand, limb pain left hand, right hand pain, Hand pain, Painful hand, Pain in hand, Pain in limb, hand, pain in the hands, limb pain hand, hands pain, |
| | Pain;hand, soft tissue pain in foot, foot pain, Burning pain in hand, Burning pain in foot, Skin symptom change, Skin change |
| Dermatitis | Dermatitides, Skin inflamed, Skin inflammation, Inflammation of Skin, skin; inflammation, inflammation; skin, Skin—Inflammation |
| Erythema/Redness | Erythema of skin, Cutaneous redness, Dermal erythema, Skin reddened, Skin red, Skin erythema, erythematous condition, unusual change in color of skin to red (erythema), redness of skin, skin redness, erythema findings, Erythematous conditions, Erythematous disorder, Red skin |
| Cardiotoxicity | Cardiac Toxicity, Cardiotoxicities, Toxicity; Cardiac, Cardiac Toxicities |
| Myocardial Infarction | Heart Attack, Infarction; Myocardial, Myocardial Infarcts, Heart Attacks, Cardiac infarction, MI, Infarction of heart, Attack coronary, Right ventricular infarction, Infarct myocardial, myocardial necrosis, infarctions myocardial, heart infarction, attack heart, attacking heart, attacks coronary, coronary attack, disorder infarction myocardial, syndrome myocardial infarction, infarction; myocardial, cardiac; infarction |
| Cardiovascular Strokes | Stroke; Cardiovascular, Strokes; Cardiovascular, Cardiovascular Stroke |
| Heart failure | Cardiac Failure, HF, Cardiac insufficiency, Weak heart, Cardiac function fail, Heart insufficiency, Failure heart, failures |
|  | heart, myocardial failure, heart weak, hearts weak, insufficiency heart, failure; heart, cardiac; failure, cardiac; insufficiency, failure; cardiac, heart; insufficiency, insufficiency; cardiac, insufficiency; heart, weak; heart, Failure;heart, Failure;cardiac, Insufficiency;cardiac, Weakness;heart, HEART FAILURE AND OTHER FUNCTIONAL DISORDERS, BLE edema, Cardiac insufficiency, Interstitial edema, Edematous, Oedema, Waterlogged, Dropsy, Hydrops, swelling due to excess fluid, excess fluid, Bilateral lower extremity edema, Exacerbation of congestive heart failure, congestive heart failure exacerbation, CHF exacerbation, Reduced left ventricular ejection fraction, Reduced ejection fraction, Reduced EF, Reduced LVEF, Reduction in EF, Cardiogenic Shock, Shock, Cardiogenic, Cardiovascular shock, Shock cardiogenic, heart shocking, heart shock, shock heart, cardiogenic; shock, shock; cardiogenic, Cardiac shock syndrome, Power failure syndrome, cardiocirculatory collapse, HF exacerbation |
| Angina | Angina Pectoris, Angor Pectoris, Stenocardia, Ischemic chest pain, Anginal syndrome, Cardiac angina, AP - Angina pectoris, Ischaemic chest pain, Anginal pain, Anginal discomfort, chest pain, Ischemic heart disease with angina, chest; pain, ischemic, pain; chest, ischemic, syndrome; anginal, anginal; |
| | syndrome, Pain;angina, ANGINAL SYNDROMES, Angina of effort, Pain in chest, Chest pain, Pain, Chest, Pains, Chest, Thorax painful, Thoracic pain, Thorax pain, Thoracalgia, Stethalgia, chest pain or discomfort, Chest discomfort, Discomfort in chest, pain thoracic, chest; pain, pain; chest, pain; thorax, thorax; pain, Pain;chest, Heart throbbing, heart irregularities, heart throb, Cardiac angina syndrome |
| Arrhythmia | Cardiac Arrhythmia, Dysrhythmia, Irregular Heartbeat, Heart Rhythm Disturbance, Cardiac Rhythm Disorder Afib, A Fib, Atrial Flutter, Auricular Flutter, A-flutter, AF, Auricular Fibrillation VF, Ventricular Fibrillation, Cardiac Arrest Due to VF, Ventricular Tachycardia (V TACH), Tachycardia;Ventricular Flutter Atrial, Fibrillation Atrial, Flutter Auricular, Heart Arrhythmia |
| Coronary Artery Vasospasms | Coronary Vasospasm, Coronary Artery Spasm, Artery Vasospasm, Coronary, Vasospasm, Coronary Artery, Coronary Vasospasms, Vasospasm, Coronary, Artery Spasm, Coronary, Coronary Artery Spasms, Spasm, Coronary Artery, Coronary spasm, Arteriospasm coronary, Coronary vascular spasm, Spasm coronary artery, artery coronary spasms, coronaries spasm, vasospasm coronary, a.coronaria; spasm, coronary; spasm, spasm; coronary, Spasm;artery;coronary |
| Dyspnea | Shortness of Breath, Breath Shortness, Difficulty breathing, Respiration difficult, DIB - Difficulty in breathing, SOB - Shortness of breath, Breathless, Dyspnoea, Abnormal breathing, Difficult to breathe, Trouble breathing, Breathing difficult, SOB (shortness of breath), tightness of breath, s.o.b. |
| Syncope | Fainting, Syncope and collapse, Blackout, Passed out, Syncopal attack, swoon, loss of consciousness; attack, Syncopal episode |
| Cardiomyopathies | Myocardial Disease, Myocardiopathies, Cardiomyopathy, Myocardiopathy, Diseases, Myocardial, Disease, Myocardial, Disorder of heart muscle, Disorder of myocardium, Myocardiodystrophy, heart muscle disease, myocardium; disease, Myocardium--Diseases |
| Myocardial Ischemia | Ischemic Heart Disease, Ischemia, Myocardial, Heart Disease, Ischemic, Disease, Ischemic Heart, IHD - Ischemic heart disease, Cardiac ischemia, Ischemia myocardial, ischemia; heart, myocardium; ischemic, ischemia; myocardial, Disease;ischaemic heart, HEART, ISCHEMIC DISEASE, Myocarditides, Myocarditis, Myocardial inflammation, Inflammation of heart muscle, myocardium; inflammation, inflammation; myocardium |

| Pericarditis | Swelling or irritation of membrane around heart, an inflammation of the membrane surrounding the heart, pericardium; inflammation, inflammation; pericardium |
|---|---|
| Endocarditis | Endocarditides, inflammation of the heart valve |
| Cardiac Arrest | Heart Arrest, Asystole, Arrest, Cardiac, Arrest, Heart, Asystoles, Cardiac standstill, Heart stops beating, Ventricular asystole, Asystolia, Ventricular asystolia, Asystolic, Ventricular arrest, Arrest cardiac, Standstill cardiac, Cardiac arrest, unspecified, cardiac asystole, stoppage; heart, arrest; cardiac, heart; arrest, heart; stoppage, ventricular; arrest, arrest; ventricular, Cardiac arrest- asystole, CA - Cardiac arrest, Cardiopulmonary Arrest, Arrest, Cardiopulmonary, Cardiorespiratory arrest, Cardio-respiratory arrest, cardiac arrest cardiorespiratory, arrest cardiopulmonary, arrest cardio respiratory, cardiorespiratory; arrest, arrest; cardiorespiratory |
| Ventricular Dysfunction | Dysfunction, Ventricular, dysfunction ventricular |
| Pericardial effusion | Effusion, Pericardial, Fluid in pericardium, Fluid around heart, Effusion pericardial, Pericardial fluid, pericardium; effusion, effusion; pericardial, effusion; pericardium |
| Heart Block | Block, Heart, Block heart, Conduction block, blocked heart, blocks heart, blockage heart, blockages heart, cardiac; block, myocardial; block, block; heart, heart; block, block; cardiac, |

|  | block; conduction, block; myocardial, conduction; block, Block;heart |
|---|---|
| Hypotension | Blood Pressure, Low, Low Blood Pressure, Hypopiesis, Blood pressure decreased, Arterial blood pressure decreased, Blood pressure drop arterial, Pressure arterial decreased, Drop of blood pressure, Fall in blood pressure, BP fell, Low BP, Blood pressure dropped, Drop in blood pressure, Blood pressure low, Lowered blood pressure, BP lowered, blood decreasing pressure, blood dropping pressure, blood drops pressure, blood falling pressure, blood falls pressure, blood lowered pressure, bp lower, decreased blood pressure, Hypotensive |
| Cardiac Tamponade | Pericardial Tamponade, Tamponade, Cardiac, Tamponade, Pericardial, Tamponade cardiac, Tamponade, tamponade; cardiac, HEART, TAMPONADE, Rose's tamponade, heart tamponade |
| Heart valve regurgitation | Valvular insufficiency, Cardiac valve insufficiency, Mitral Valve Insufficiency, MITRAL REGURGITATION, Mitral Insufficiency, Mitral Incompetence, Mitral Valve Incompetence, Valve Insufficiency, Mitral, Insufficiency, Mitral Valve, Regurgitation, Mitral, Regurgitation, Mitral Valve, Insufficiency, Mitral, MI - Mitral incompetence, MR - Mitral regurgitation, Regurgitation of left atrioventricular valve, Ventriculo-atrial regurgitation |
| Valvular complications | Tricuspid Valve Insufficiency, Tricuspid Valve Regurgitation, Tricuspid Incompetence, Tricuspid Regurgitation, Insufficiency, Tricuspid Valve, Valve Insufficiency, Tricuspid, Regurgitation, Tricuspid, Regurgitation, Tricuspid Valve, Valve Incompetence, Tricuspid, Incompetence, Tricuspid, Incompetence, Tricuspid Valve, TR - Tricuspid regurgitation, TI - Tricuspid incompetence, Regurgitation of right atrioventricular valve, heart valve disorder atrioventricular right leaflet, abnormality regurgitation, Aortic Valve Insufficiency, Aortic Valve Incompetence, Regurgitation, Aortic Valve, Aortic Regurgitation, Insufficiency, Aortic Valve, Incompetence, Aortic, Incompetence, Aortic Valve, Regurgitation, Aortic, Aortic valve regurgitation, Aortic insufficiency, AI - Aortic incompetence, AR - Aortic regurgitation, Aortic incompetence, Aortic (valve) insufficiency |
| Cyanosis | Cyanoses, unusual change in color of skin to blue, skin cyanosis |